\documentclass[10pt,twocolumn,letterpaper]{article}

\usepackage{iccv}
\usepackage{times}
\usepackage{epsfig}
\usepackage{graphicx}
\usepackage{amsmath}
\usepackage{amssymb}

\usepackage{booktabs}       
\usepackage{amsthm}
\usepackage{amsfonts}       
\usepackage[utf8]{inputenc} 
\usepackage[T1]{fontenc}    
\usepackage[ruled,vlined]{algorithm2e}
\usepackage{nicefrac}       
\usepackage{microtype}      
\usepackage{multirow}
\usepackage{multicol}
\usepackage{makecell}

\newtheorem{proposition}{Proposition}

\newtheorem{innercustompro}{Proposition}
\newenvironment{custompro}[1]
  {\renewcommand\theinnercustompro{#1}\innercustompro}
  {\endinnercustompro}

\graphicspath{ {./figure/} }
\usepackage{float}
\usepackage{cellspace}
\usepackage{wrapfig}
\usepackage{layouts}
\usepackage{subcaption}
\usepackage{caption}
\usepackage{bbm}
\usepackage{nameref}
\usepackage{comment}
\newcommand{\specialcell}[2][c]{%
  \begin{tabular}[#1]{@{}c@{}}#2\end{tabular}}
  
\restylefloat{figure}
\usepackage{wrapfig}
\usepackage[accsupp]{axessibility}

\usepackage[pagebackref=true,breaklinks=true,letterpaper=true,colorlinks,bookmarks=false]{hyperref}

\usepackage{cleveref}

\iccvfinalcopy 

\def\iccvPaperID{9002} 

\ificcvfinal\fi

\begin{document}

\title{Reliably fast adversarial training via latent adversarial perturbation}

\author{Geon Yeong Park \qquad Sang Wan Lee \\
KAIST\\
Daejeon, South Korea\\
{\tt\small \{pky3436, sangwan\}@kaist.ac.kr}}


\maketitle
\ificcvfinal\thispagestyle{empty}\fi

\begin{abstract}
   While multi-step adversarial training is widely popular as an effective defense method against strong adversarial attacks, its computational cost is notoriously expensive, compared to standard training. Several single-step adversarial training methods have been proposed to mitigate the above-mentioned overhead cost; however, their performance is not sufficiently reliable depending on the optimization setting. To overcome such limitations, we deviate from the existing input-space-based adversarial training regime and propose a single-step latent adversarial training method (\textbf{SLAT}), which leverages the gradients of latent representation as the latent adversarial perturbation. We demonstrate that the \(\ell_1\) norm of feature gradients is implicitly regularized through the adopted latent perturbation, thereby recovering local linearity and ensuring reliable performance, compared to the existing single-step adversarial training methods. Because latent perturbation is based on the gradients of the latent representations which can be obtained for free in the process of input gradients computation, the proposed method costs roughly the same time as the fast gradient sign method. Experiment results demonstrate that the proposed method, despite its structural simplicity, outperforms state-of-the-art accelerated adversarial training methods.
\end{abstract}

\section{Introduction}
\label{sec:intro}
Although several studies have suggested the use of deep learning methods to solve challenging tasks, adversarial vulnerability \cite{szegedy2013intriguing} is one of the remaining major challenges while employing deep learning to safety-critical applications. Adversarial training (AT) approaches aim to mitigate the problem by training the model on generated adversarial examples, i.e. the sample corrupted with human-imperceptible noise which can fool the state-of-the-art deep neural networks. Although PGD AT \cite{madry2017towards} is one of the most effective training methods, it consumes a considerable training time because it relies on multiple projected gradient descent steps to generate the adversaries. The AT based on Fast Gradient Sign Method (FGSM; \cite{goodfellow2014explaining}) reduces the training time; however, recent works \cite{madry2017towards, tramer2017space, tramer2017ensemble} have identified the FGSM's vulnerability to the sophisticated adversaries.

\begin{figure*}[htbp]
\centering
\begin{subfigure}[c]{0.46\textwidth}
\includegraphics[width=\textwidth]{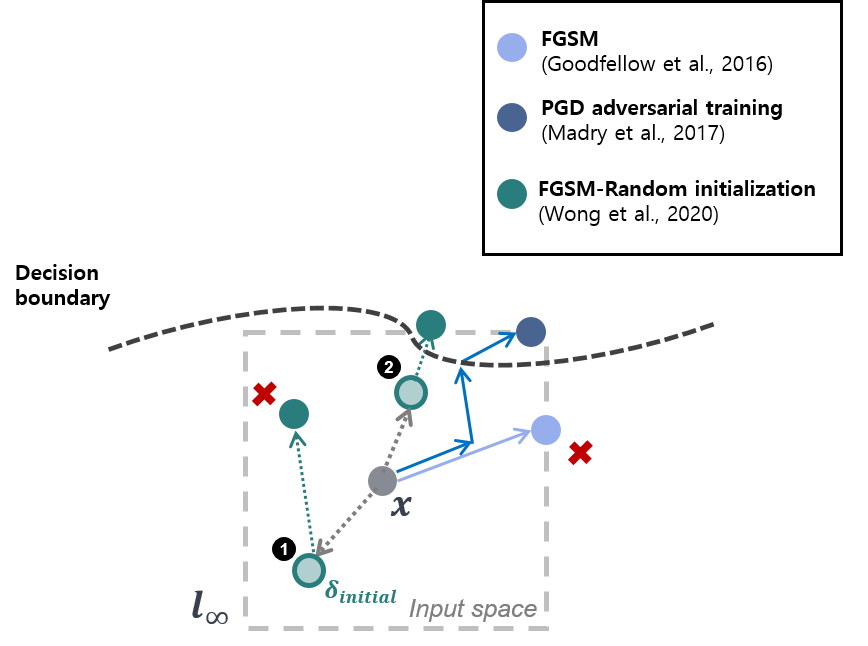} 
\caption{Existing approaches}\label{fig:concept_prev_AT}
\end{subfigure}
\begin{subfigure}[c]{0.3\textwidth}
\includegraphics[width=\textwidth]{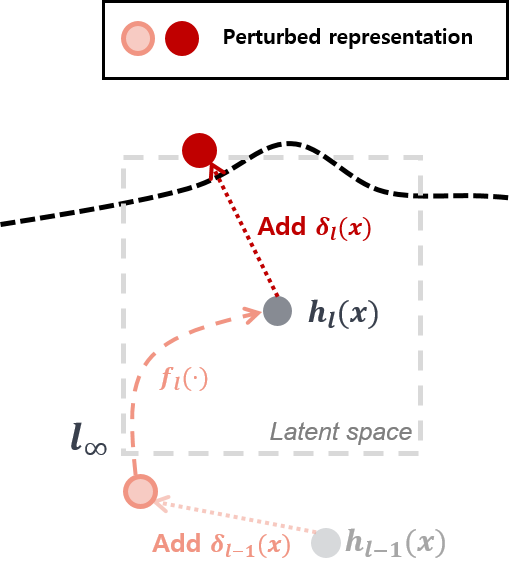}
\caption{\textbf{SLAT}}\label{fig:concept_SSH_AT}
\end{subfigure}
\caption{Visual illustration of the conceptual difference between existing and proposed approach. \textbf{(a)} FGSM may fail to generate the appropriate adversary because it approximates the solution of inner maximization problem with a single gradient step. While PGD-based AT may generate relatively more desirable adversary, it takes multiple iterations per sample to solve the inner maximization which is computationally expensive. Uniform random initialization \cite{wong2020fast} contributes to improving the performance of FGSM; however, the success of such initialization is not fundamentally justified. \textbf{(b)} Our proposed method mitigates the suggested problems by introducing latent adversarial perturbations in parallel.} \label{fig:concept}
\end{figure*}

The trade-off between adversarial robustness and computational cost has facilitated the development of accelerated and trustworthy AT methods. Shafahi et al. \cite{shafahi2019adversarial} significantly reduced the computational burden by presenting a \textit{free} AT method that updates both model parameters and adversarial perturbation through a single shared backward propagation. Wong et al. \cite{wong2020fast} proposed a \textit{fast} adversarial training based on the discovery that a slight modification in the FGSM training method such as random initialization allows it to achieve an adversarial robustness on par with PGD AT. They also discovered the \textit{catastrophic overfitting} problem of FGSM AT, wherein the model suddenly loses its robustness during training within an epoch.

Although substantial technical advances have been made with regard to the above-mentioned methods, recent works have reported that such approaches are not sufficiently reliable. Andriushchenko et al. \cite{andriushchenko2020understanding} demonstrated that fast adversarial training still suffers from the catastrophic overfitting, owing to the deteriorated local linearity of neural networks. Kim et al. \cite{kim2020understanding} found that the fast adversarial training suddenly loses its robustness and eventually collapses when a simple multi-step learning rate schedule is used. Li et al. \cite{li2020towards} reported that although fast adversarial training may recover quickly, it still temporally exhibits catastrophic overfitting.

This remaining problem in AT motivates us to explore novel ways to improve the reliability of single-step AT, without bearing a considerable training time. In this study, we demonstrate that the single-step latent adversarial training (\textbf{SLAT}) with the latent adversarial perturbation operates more effectively and reliably compared to the other single-step adversarial training variants. While many of existing adversarial training methods require multiple gradient computations which is inevitably time-consuming, we exploit the gradients of latent representations from multiple layers in \textit{parallel} for the synergistic generation of adversary. Note that the gradients of latent representations can be obtained for \textit{free} in the process of computation of input gradients.

To the best of our knowledge, our work is the first to reduce the computational cost of AT by deviating from the input-space based frameworks and injecting latent adversarial perturbations. The contribution of this study is summarized as follows. First, we propose that the local linearity of neural networks can be regularized without demanding cost of training time by adopting latent adversarial perturbation, unlike the gradient alignment (GA) regularization \cite{andriushchenko2020understanding}, which is three times slower compared to FGSM training. In particular, we demonstrates that the \(\ell_1\) norm of feature gradients is implicitly regularized by introducing latent adversarial perturbation, which closes the gap between the loss function of neural networks and its first-order approximation. As the latent adversarial perturbation is adopted across multiple latent layers, the synergistic regularization effect can be expected. Second, we demonstrate that \textbf{SLAT} outperforms the state-of-the-art accelerated adversarial training methods, while achieving performance comparable to PGD AT.

\section{Related works} 
\label{sec:related_works}
Adversarial training has been improved with the help of adversarial attack algorithms. Goodfellow et al. \cite{goodfellow2014explaining} proposed FGSM to enable rapid generation of adversarial examples through a single-step gradient update. Based on these developments, madry et al. \cite{madry2017towards} generated a stronger adversary through projected gradient descent (PGD). PGD-based AT has been recognized as a more effective defense method than others, such as provable defenses \cite{wong2018provable, zhang2019towards, cohen2019certified}, label smoothing \cite{shafahi2019label}, mix-up \cite{zhang2017mixup}, and Jacobian regularization \cite{jakubovitz2018improving}. However, the complexity overhead of generating the adversary significantly limits the scalability of the PGD-based AT method.

As introduced in Section \ref{sec:intro}, several methods \cite{shafahi2019adversarial, wong2020fast} have been proposed to improve the efficiency of AT. Zhang et al. \cite{zhang2019you} analyzed AT from the perspective of a differential game and merged the inner loop of the PGD attack and the gradient update of model parameters. Vivek et al. \cite{vivek2020single} experimentally found that the adversarial robustness of FGSM AT is improved via application of dropouts to all non-linear layers. Kim et al. \cite{kim2020understanding} demonstrated that the characteristic of FGSM AT, which uses only adversary with the maximum perturbation, leads to the decision boundary distortion and therefore proposed an ad-hoc method to determine an appropriate step size. While most of the proposed methods heavily rely on the input perturbation, we explore the possibility for efficient AT using latent adversarial perturbations.

Although several remarkable works \cite{xie2019feature, wang2020high} spark interest regarding the importance of latent representations in adversarial robustness, only a few works have directly leveraged the latent adversarial perturbations for AT. Sankaranarayanan et al. \cite{sankaranarayanan2018regularizing} used the gradients of latent representations computed from the preceding mini-batch to approximate the solution of inner-maximization. Although the proposed method has yielded modest improvements in terms of the adversarial robustness with respect to FGSM-based AT, they do not necessarily mean that a latent adversarial perturbation from a gradient of the preceding mini-batch is optimal. While \cite{singh2019harnessing} suggested latent adversarial training for further fine-tuning of the adversarially trained model, it is contingent on performance of multi-step PGD AT. On the contrary, our approach focusses on the reduction of computational costs of AT via latent adversarial perturbations.

\section{Latent adversarial perturbation}
\label{sec:latent}
We begin with formulating the generalized min-max adversarial training objective in terms of latent adversarial perturbation. Let \((x \in \mathcal{X}, y \in \mathcal{Y}) \sim D\) be the pair of sample and label instance generated from the distribution \(D\), given sample space \(\cal{X}\) and label space \(\cal{Y}\). We represent \(f_l(\cdot)\) as the function defined by the \(l\)-th layer, and \(h_l(x)\) as the latent-representation vector given sample \(x\), where \(h_0(x) \triangleq x\). Precisely, \(h_l(x) = f_l(f_{l-1}(\dots(f_1(x))))=f_l(h_{l-1}(x))\), where \(f_l(h_{l-1}(x)) = \phi(W_l h_{l-1}(x)+b_l)\), given the \(l\)-th weight matrix \(W_l\), bias vector \(b_l\), and the activation function \(\phi(\cdot)\). Note that the latent adversarial perturbation is not considered yet. We denote the \(L\)-layer neural networks as a function \(f_{\pmb{\theta}}:\cal{X} \rightarrow \cal{Y}\), parameterized by \(\pmb{\theta} = \{W_1, \dots, W_L, b_1, \dots, b_L\}\): \(f_{\pmb{\theta}}(x)=f_L(f_{L-1}(\dots (f_1(x))))\). For simplicity, let \(f_{m:n}(x) = f_n(f_{n-1}(\dots(f_m(h_{m-1}(x)))))\), for any \(1 \leq m < n \leq L\).

In this study, we investigate the benefits of latent adversarial perturbation, which is built based on the gradients of latent representations. Let \(K \subseteq \{0, \dots, L-1\}\) be the subset of layer indexes injected with adversarial perturbation, where the \(0\)th layer represents the input layer. We denote the adversarial perturbation given \(x\) for the \(k\)-th layer as  \(\delta_k(x)\), where \(k \in K\), and the set of adversarial perturbations is \(\pmb{\delta} = \{\delta_k(x)\}_{\forall k \in K} \). 

To examine the marginalized effect of latent adversarial perturbations, we define the \textit{accumulated} perturbation in layer \(L-1\) as \(\hat{\delta}_{\ell-1}(x)\), where \(\hat{\delta}_{\ell-1}(x)\) originates from the forward propagation of \(\pmb{\delta}\). Note that we virtually introduced the accumulated perturbation for the sake of the analysis. In practice, each latent adversarial perturbation is applied layer-wise. Thus, \(h_{l+1}(x) = f_l(h_l(x) + \delta_l(x)), \forall l \in K\). 

The optimal set of parameters \(\pmb{\theta}^*\) and adversarial perturbations \(\pmb{\delta}^*\) can be obtained by solving the following min-max problem:
\begin{equation}
\label{eq:adversarial_noise_objective}
\begin{split}
    &\pmb{\theta^*}, \pmb{\delta^*} = \\
    &\arg \min_{\pmb{\theta}} \mathbb{E}_{(x,y) \sim D} \max_{\pmb{\delta}} \bigg[ {\mathcal{L}} \Big(f_{L}\big(h_{L-1}(x) + \hat{\delta}_{\ell-1}(x) \big), y\Big) \bigg].
\end{split}
\end{equation}

While it is difficult to obtain the accumulated perturbation \(\hat{\delta}_{\ell-1}(x)\) in a closed form given highly nonlinear function \(f_{\pmb{\theta}}(\cdot)\), we can approximate it by making a reasonable assumption that \(\delta_k(x)\) is sufficiently small. The approximation is based on \cite{camuto2020explicit} which examined the effect of Gaussian Noise Injection (GNI) into multiple latent layers. The adversarial perturbation accumulated on the layer \(L-1\) can be expressed as follows:
\begin{proposition}
\label{pro:accum_perturb}
Consider an \(L\) layer neural network, with the latent adversarial perturbations \(\delta_k(x)\) being applied at each layer \(k \in K\). Assuming the Hessians, of the form \(\nabla^2 h_l(x)|_{h_m(x)}\) where \(l, m\) are the index over layers, are finite. Then the perturbation accumulated at the layer \(L-1\), \(\hat{\delta}_{\ell-1}(x)\), is approximated by:
\begin{equation}
\hat{\delta}_{\ell-1}(x) = \sum_{k \in K} \mathbf{J}_k(x) \delta_k(x) + O(\gamma),
\end{equation}

where \(\mathbf{J}_k(x) \in \mathbb{R}^{N_{L-1} \times N_k}\) represents each layer's Jacobian; \(\mathbf{J}_k(x)_{i,j} = \frac{\partial h_{L-1}(x)_i}{\partial h_k(x)_j}\), given the number of neurons in layer \(L-1\) and \(k\) as \(N_{L-1}\) and \(N_k\), respectively. \(O(\gamma)\) represents higher order terms in \(\pmb{\delta}\) that tend to zero in the limit of small perturbation.
\end{proposition}
The detailed proof is provided in the supplementary material. Based on the framework (\ref{eq:adversarial_noise_objective}) and Proposition \ref{pro:accum_perturb}, we provide the details of the proposed latent adversarial perturbation. By neglecting the higher order terms in Proposition \ref{pro:accum_perturb}, the linear approximation of the loss function \({\mathcal{L}} \Big(f_{L}\big(h_{L-1}(x) + \hat{\delta}_{\ell-1}(x) \big), y\Big)\) is as follows:
\begin{equation}
\label{eq:obj_first_order_approx}
\begin{split}
&{\mathcal{L}} \Big(f_{L}\big(h_{L-1}(x) + \hat{\delta}_{\ell-1}(x) \big), y\Big) \\
&\approx {\mathcal{L}} \Big(f_{L}\big(h_{L-1}(x), y\big)\Big) + \\
& \qquad \nabla_{{h_{L-1}(x)}}  {\mathcal{L}} \Big(f_{L}\big(h_{L-1}(x)\big), y \Big)^T \hat{\delta}_{\ell-1}(x) \\ 
&= {\mathcal{L}} \Big(f_{L}\big(h_{L-1}(x), y\big)\Big) + \\
& \qquad \nabla_{{h_{L-1}(x)}}  {\mathcal{L}} \Big(f_{L}\big(h_{L-1}(x)\big), y \Big)^T \sum_{k \in K} \mathbf{J}_k(x) \delta_k(x).
\end{split}
\end{equation}

Then, we approximate the solution of the inner maximization problem in (\ref{eq:adversarial_noise_objective}), as similarly done in FGSM:
\begin{equation}
\label{eq:latent_adv_solution}
\begin{split}
\delta_k(x) &= \eta_k \cdot sign \bigg(\mathbf{J}_k(x)^T \nabla_{h_{L-1}(x)} {\mathcal{L}} \big(f_{L}(h_L(x)), y\big) \bigg) \\
&= \eta_k \cdot sign \bigg(\nabla_{h_{k}(x)} {\mathcal{L}} \big(f_{\pmb{\theta}}(x), y \big) \bigg), \forall k \in K,
\end{split}
\end{equation}
where \(\eta_k\) is the step size for the \(k\)-th layer. Then, the explicit regularizer of latent adversarial perturbation in (\ref{eq:obj_first_order_approx}) is derived as follows:
\begin{equation}
\label{eq:adversarial_noise_distribution}
\begin{split}
{\mathcal{L}}_{latent} &= \sum_{k \in K} \eta_k \cdot sign\bigg(\nabla_{h_{k}(x)} {\mathcal{L}}(f_{\pmb{\theta}}(x), y) \bigg) \\
& \qquad \qquad \qquad \qquad \circ \nabla_{h_{k}(x)} {\mathcal{L}}(f_{\pmb{\theta}}(x), y) \\
&= \sum_{k \in K} \eta_k \cdot ||\nabla_{h_{k}(x)} {\mathcal{L}}(f_{\pmb{\theta}}(x), y)||_1,
\end{split}
\end{equation}
where \(\circ\) represents dot product. 

It affords us the theoretical insights that the latent adversarial perturbation leads to the implicit regularization of \(\ell_1\) norm of the feature gradients. Although the uses of input gradient regularization in adversarial defense \cite{ross2018improving} or explainable machine learning \cite{ross2017neural, smilkov2017smoothgrad} have been widely recognized, the effects of feature gradient regularization on adversarial robustness have been poorly understood. To address this, we establish a connection between the feature gradient regularization and local linearity. For a better linear approximation, the linear approximation error \(R_k(x)\) should be constrained: 
\begin{equation}
\label{eq:linear_approx_error}
\begin{split}
\Big| &{\mathcal{L}}\Big(f_{k:L} \big(h_{k-1}(x)+\epsilon \big), y\Big) - {\mathcal{L}}\Big(f_{k:L} \big(h_{k-1}(x) \big), y\Big) \\
&- \Big\langle \nabla_{h_{k-1}(x)} {\mathcal{L}}\Big(f_{k:L} \big(h_{k-1}(x) \big), y\Big), \epsilon \Big\rangle \Big|,
\end{split}
\end{equation}
where \(\epsilon\) is an arbitrary perturbation with sufficiently small size, and \(R_k(x)\) is defined with the function \(f_{k:L}(\cdot)\) for an arbitrary \(k \in K\). Note that \(R_k(x)\) includes the second-order term \(\big \langle \epsilon, H_k(x)\epsilon \big \rangle\) where \(H_k(x)=\nabla^2_{h_{k-1}(x)} {\mathcal{L}}\big(f_{k:L}(h_{k-1}(x)), y\big)\). Let \(\nabla_{k-1}\) be a shorthand for \(\nabla_{h_{k-1}(x)} {\mathcal{L}}\big(f_{k:L}(h_{k-1}(x)), y\big)\). By the low-rank approximation of the Hessian matrix \cite{martens2012estimating}, let \(H_k(x) \approx \nabla_{k-1} \nabla_{k-1}^T\). Then, the upper bound of the second-order term is as follows:
\begin{equation}
\label{eq:upper_bound_second_term}
\begin{split}
\langle \epsilon, H_k(x)\epsilon \big \rangle &\approx |\langle \epsilon, \nabla_{k-1} \rangle|^2 \\
&\leq ||\epsilon||^2_2 ||\nabla_{k-1}||^2_2 \\
&\leq ||\epsilon||^2_2 ||\nabla_{k-1}||^2_1,
\end{split}
\end{equation}
since \(||x||_p \geq ||x||_q\) for \(0<p<q, \forall x \in \mathbb{R}^n\). Thus the regularization of \(\ell_1\) norm of feature gradients may result in a better linear approximation of the loss function. This eventually contributes to improving the reliability of FGSM which relies heavily on linear approximation of the loss function. 

We further foster a close collaboration between feature gradient regularization and the minimization of adversarial loss. Inspired from \cite{simon2019first}, the small variation in the loss \(\bigtriangleup {\mathcal{L}}_k\) caused by the latent adversarial perturbation \(\delta_k(x)\) is as follows:
\begin{equation}
\label{eq:adversarial_noise_distribution}
\begin{split}
&\bigtriangleup {\mathcal{L}}_k \\
&= \max_{||\delta_k(x)|| \leq \eta_k} \Big| {\mathcal{L}}\Big(f_{k+1:L} \big(h_k(x)+\delta_k(x) \big), y\Big) \\
& \qquad \qquad \qquad \qquad - {\mathcal{L}}\Big(f_{k+1:L} \big(h_k(x) \big), y\Big) \Big| \\
&\approx \max_{||\delta_k(x)|| \leq \eta_k} \Big| \Big\langle \nabla_{h_k(x)} {\mathcal{L}}\Big(f_{k+1:L} \big(h_{k}(x) \big), y \Big), \delta_k(x) \Big \rangle \Big| \\
&= \eta_k \Big|\Big|\nabla_{h_k(x)} {\mathcal{L}}\Big(f_{k+1:L} \big(h_k(x) \big), y \Big) \Big|\Big|_{*},
\end{split}
\end{equation}
where \(\eta_k\) is the allowed step size for the perturbation \(\delta_k(x)\). The last equality comes from the definition of the dual norm \(||\cdot||_*\) of \(||\cdot||\). Thus the regularization of \(\ell_1\) norm of feature gradients is closely related to minimizing the adversarial loss stems from \(\delta_k(x)\) with limited \(\ell_\infty\) norm.

\textbf{Advantages over other local linearity regularization methods.}  Existing works have been proposed to theoretically investigate the underlying principles of adversarial training with respect to local linearity. Moosavi et al. \cite{moosavi2019robustness} demonstrated that adversarial training increases adversarial robustness by decreasing curvature and proposed a new curvature regularizer based on the finite difference approximation of Hessian. While the proposed regularizer improves the adversarial robustness with affordable computational costs, it requires pretrained networks for fine-tuning. Moreover, the reported standard and adversarial accuracy was lower than that of our proposed method given the same base network architecture (Section \ref{sec:experiments}). 

A few follow-up studies have attempted to regularize linear approximation errors in an optimal manner. Qin et al. \cite{qin2019adversarial} explores the worst-case perturbation that maximizes the linear approximation error of the loss function via multiple steps of PGD in order to regularize local linearity and penalize gradient obfuscation \cite{athalye2018obfuscated}. Tsiligkaridis et al. \cite{tsiligkaridis2020frank} proposed a novel adversarial training method based on the Frank-Wolfe algorithm \cite{jaggi2013revisiting} to decrease the directional variation of loss gradients. Compared to existing works that inevitably rely on the multi-step gradient updates for regularizing local linearity, the above latent adversarial perturbation (\ref{eq:latent_adv_solution}) enables us to seamlessly regularize the local linearity of sub-networks in parallel with affordable computational costs.

\begin{algorithm}[h]
\DontPrintSemicolon
\caption{Single-step Latent Adversarial Training method (\textbf{SLAT})}
\label{algo:adversarial_training}
{\bfseries Input:} Training iteration \(T\), Number of samples \(N\), Number of layers \(L\), Training set \(\mathcal{D}= \{(x_i, y_i)\}_{i=1}^{N}\), Subset of layer indexes \(K\), Layer-wise step size \(\eta_k\) \;

{\bfseries Output:} Adversarially robust network \(f_{\pmb{\theta}}\) \;

\For{\(t \gets 1\) \textbf{to} \(T\)} {
    \For{\(i \gets 1\) \textbf{to} \(N\)} {
        \For{\(k \in K\)} {
            // Compute latent adversarial perturbations \;
            \(\delta_k(x_i) = \eta_k \cdot sign \bigg(\nabla_{h_{k}(x_i)} {\mathcal{L}}\big(f_{\pmb{\theta}}(x_i), y_i \big) \bigg)\) 
        }    
        
        \For{\(l \in \{0, \dots, L-2\}\)} { 
            \If{\(l\in K\)} {
            // Propagate adversarial perturbations forward \;
                \( h_{l+1}(x_i) = f_{l+1}(h_l(x_i) + \delta_l(x_i)) \) \;
            }
            \Else {
                \(h_{l+1}(x_i) = f_{l+1}(h_l(x_i)) \) \;
            }
        }
        Optimize \(\pmb{\theta}\) by the objective \({\mathcal{L}}(f_L(h_{L-1}(x_i)), y_i)\) using gradient descent. \;
    }
}
\end{algorithm}

\section{Experiments}
\label{sec:experiments}

\begin{table*}[htbp]
\caption{Standard and robust accuracies (\(\%\)) on CIFAR-10, CIFAR-100, and Tiny ImageNet datasets.} 
\label{table:CIFAR-10}
\centering
\begin{tabular}{c c c c c c}
\toprule
{} & Method & {Standard} & {PGD-50-10} & {AutoAttack} & {Training time (min)} \\

\midrule

\parbox[t]{2mm}{\multirow{8}{*}{\rotatebox[origin=c]{90}{CIFAR-10}}} & 
PGD-7 & {84.86\(\pm\)0.16} & {51.63\(\pm\)0.13} & {48.65\(\pm\)0.08}  & {383.2} \\
& FGSM-GA & {82.88\(\pm\)0.01} & {48.90\(\pm\)0.37} & {46.22\(\pm\)0.30} & {297.9} \\
\cmidrule{2-6}

&YOPO-5-3 & {82.35\(\pm\)1.78} & {34.23\(\pm\)3.61} & {32.79\(\pm\)3.65} & {62.5} \\
&Free-AT (\(m=8\)) & {76.57\(\pm\)0.19} & {44.15\(\pm\)0.30} & {41.02\(\pm\)0.20} & {119.4} \\
&FGSM & {87.42\(\pm\)1.08} & {0.01\(\pm\)0.01} & {0.00\(\pm\)0.00}  & {100.5} \\
&FGSM-RS & {90.76\(\pm\)6.36} & {3.90\(\pm\)4.06} & {0.44\(\pm\)0.50} & {99.7} \\
&FGSM-CKPT (\(c=3\)) & {89.32\(\pm\)0.10} & {40.83\(\pm\)0.36} & {39.38\(\pm\)0.24} & {121.4} \\
&\textbf{SLAT} & {85.91\(\pm\)0.31} & {47.06\(\pm\)0.03} & {44.62\(\pm\)0.11} & {104.6} \\

\midrule

\parbox[t]{2mm}{\multirow{8}{*}{\rotatebox[origin=c]{90}{CIFAR-100}}} & 

PGD-7 & {59.59\(\pm\)0.17} & {29.58\(\pm\)0.24} & {26.00\(\pm\)0.20} & {392.1} \\
& FGSM-GA & {58.63\(\pm\)0.17} & {27.53\(\pm\)0.10} & {24.07\(\pm\)0.15} & {240.5} \\

\cmidrule{2-6}
&YOPO-5-3 & {51.45\(\pm\)7.33} & {15.23\(\pm\)2.01} & {13.94\(\pm\)1.82} & {65.0} \\
&Free-AT (\(m=8\)) & {48.02\(\pm\)0.29} & {22.40\(\pm\)0.19} & {18.67\(\pm\)0.03} & {117.1} \\
&FGSM & {61.96\(\pm\)2.17} & {0.00\(\pm\)0.00} & {0.00\(\pm\)0.00} & {99.9} \\
&FGSM-RS & {50.96\(\pm\)4.57} & {0.00\(\pm\)0.00} & {0.00\(\pm\)0.00} & {100.9}\\
&FGSM-CKPT (\(c=3\)) & {73.53\(\pm\)0.65} & {0.66\(\pm\)0.60} & {0.09\(\pm\)0.09} & {101.5}\\
&\textbf{SLAT} & {59.56\(\pm\)0.50} & {26.26\(\pm\)0.47} & {23.02\(\pm\)0.14} & {101.7}\\
\midrule

\parbox[t]{2mm}{\multirow{8}{*}{\rotatebox[origin=c]{90}{Tiny ImageNet}}} & 
PGD-7 & {48.92\(\pm\)0.43} & {23.05\(\pm\)0.35} & {18.78\(\pm\)0.14}  & {3098.3} \\
& FGSM-GA & {48.73\(\pm\)0.14} & {22.62\(\pm\)0.11} & {18.34\(\pm\)0.07} & {2032.2} \\
\cmidrule{2-6}

&YOPO-5-3 & {51.45\(\pm\)6.01} & {15.08\(\pm\)1.78} & {13.94\(\pm\)1.61} & {511.5} \\
&Free-AT (\(m=8\)) & {22.40\(\pm\)0.17} & {9.05\(\pm\)0.08} & {6.06\(\pm\)0.18} & {911.6} \\
&FGSM & {36.47\(\pm\)11.75} & {8.68\(\pm\)12.27} & {6.63\(\pm\)9.38}  & {779.8} \\
&FGSM-RS & {42.13\(\pm\)14.98} & {10.32\(\pm\)11.93} & {8.41\(\pm\)9.73} & {787.5} \\
&FGSM-CKPT (\(c=3\)) & {61.64\(\pm\)2.24} & {5.91\(\pm\)6.68} & {5.26\(\pm\)5.98} & {753.0} \\
&\textbf{SLAT} & {48.77\(\pm\)0.25} & {20.21\(\pm\)0.16} & {16.38\(\pm\)0.16} & {785.5} \\

\bottomrule
\end{tabular}
\end{table*}

\subsection{Experiments on Toy Dataset}

\begin{figure}[htbp]
\centering
\begin{subfigure}[c]{0.15\textwidth}
\includegraphics[width=\textwidth]{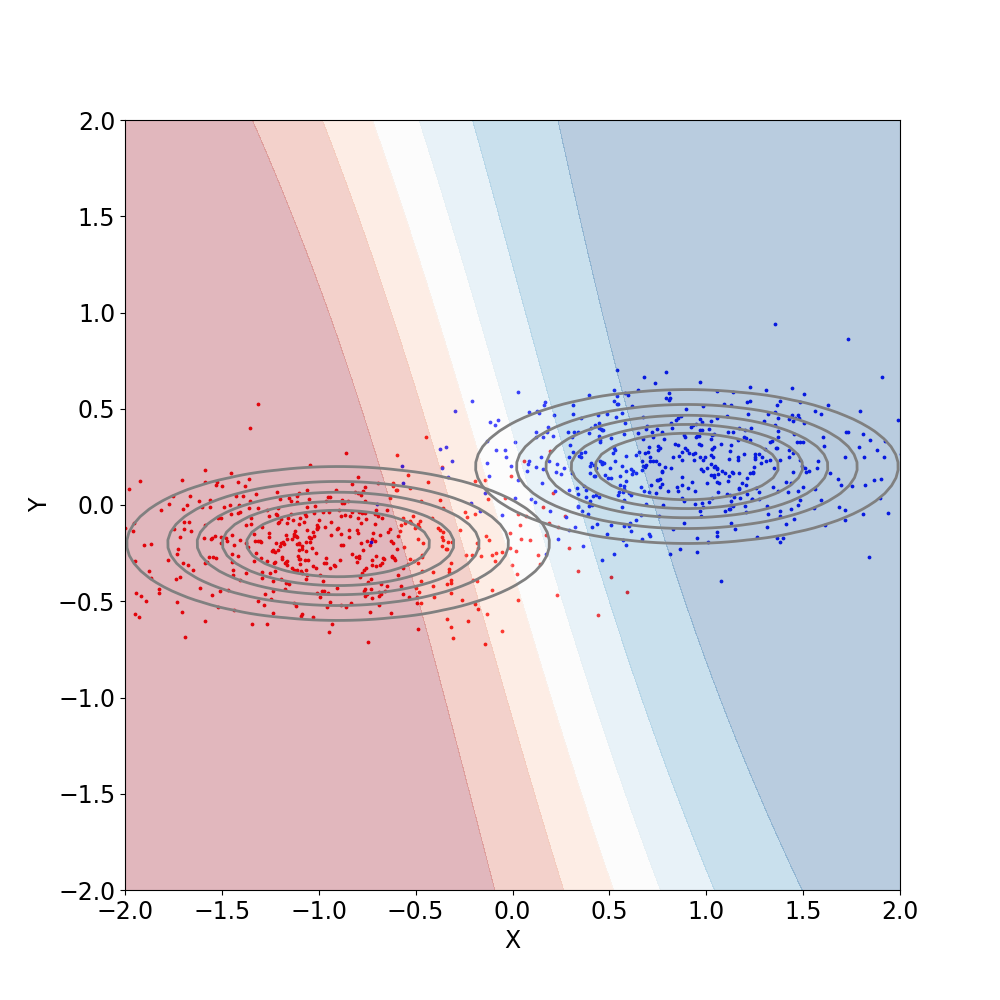} 
\caption{Standard}\label{fig:bdry_standard}
\end{subfigure}
\begin{subfigure}[c]{0.15\textwidth}
\includegraphics[width=\textwidth]{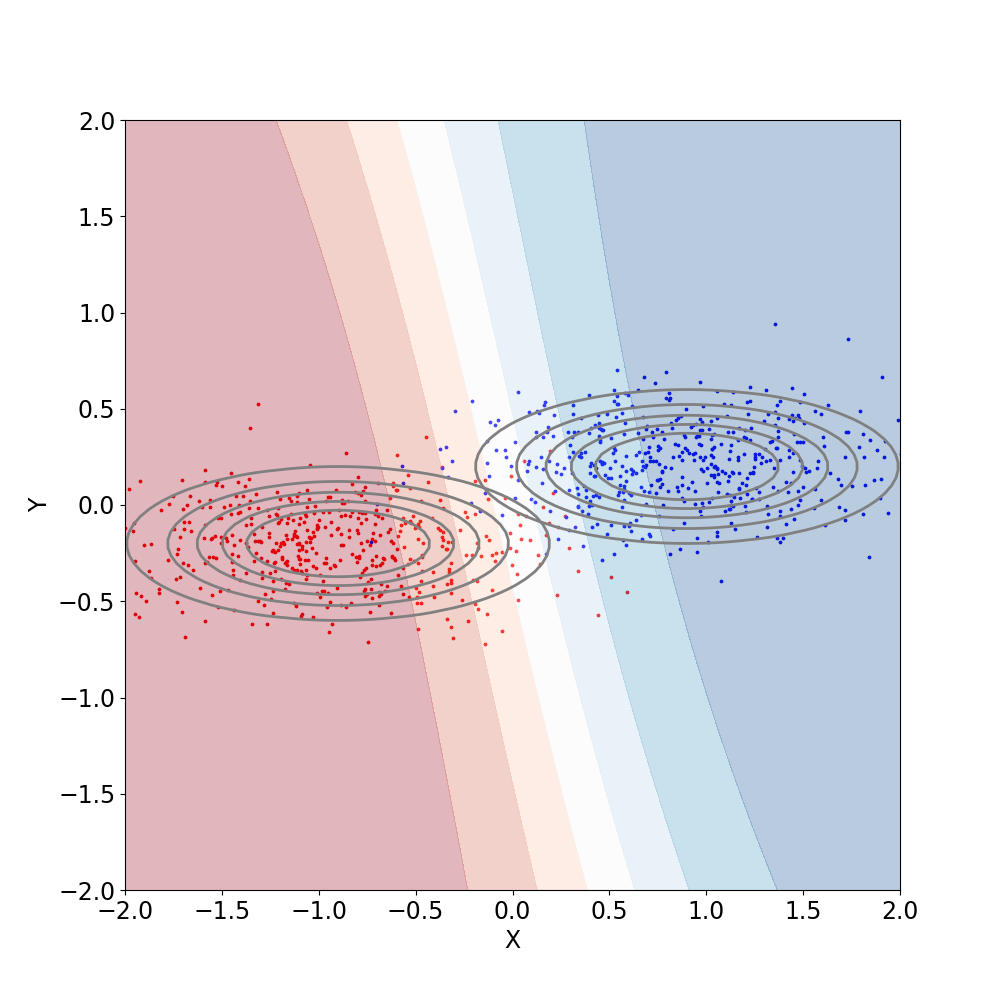}
\caption{FGSM-AT}\label{fig:bdry_FGSM}
\end{subfigure}
\begin{subfigure}[c]{0.15\textwidth}
\includegraphics[width=\textwidth]{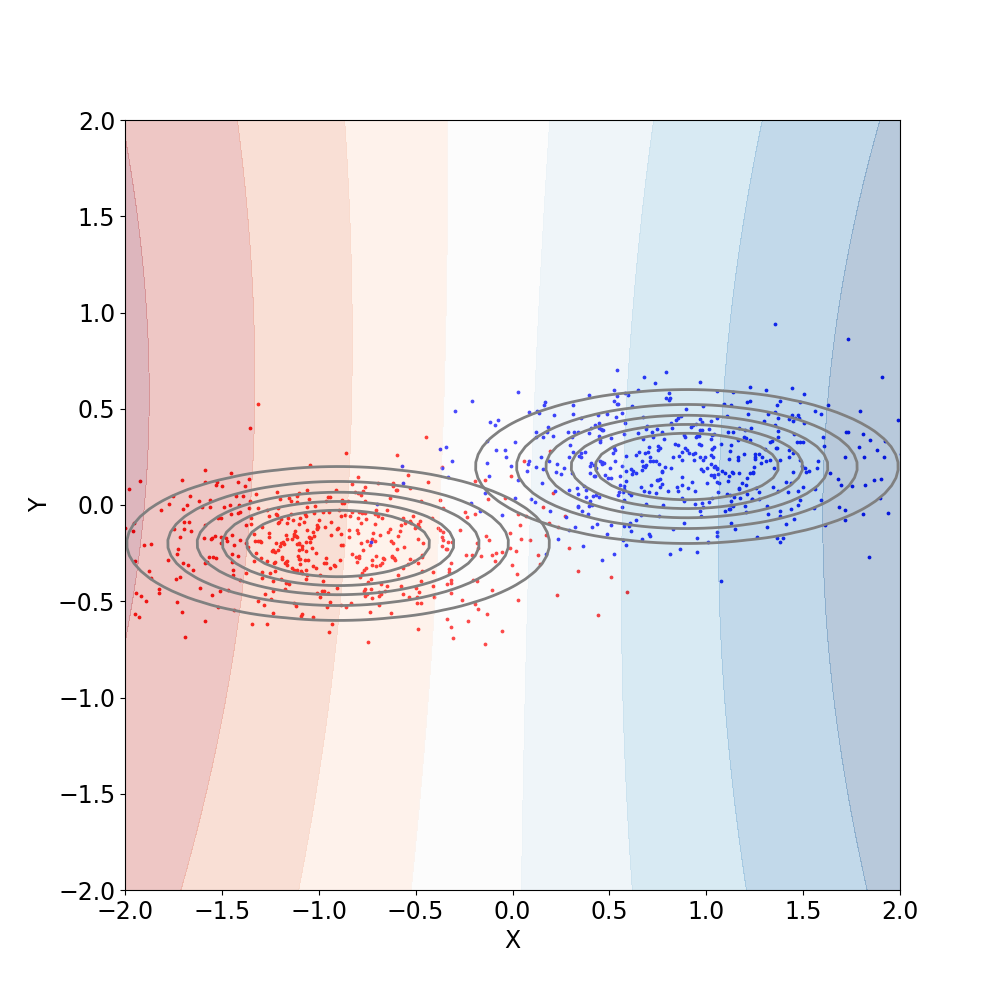}  
\caption{\textbf{SLAT}}\label{fig:bdry_advGNI}
\end{subfigure}
\caption{Decision boundary of binary classifiers. Best viewed in color.} \label{fig:toy}
\end{figure}

To conceptually clarify the effects of latent adversarial perturbation, we observed the behavior of the classifier on a simple binary classification problem. We generate samples \((x,y)\) from 2D gaussian distributions, \(\mathcal{N}_1(\mu, \Sigma)\), and \(\mathcal{N}_2(-\mu, \Sigma)\). Ilyas et al. \cite{ilyas2019adversarial} proposed that the adversarial samples stem from useful but non-robust features which are informative for classification in standard setting but vulnerable to adversarial attacks. Inspired by \cite{ilyas2019adversarial}, we intentionally compose the input features as robust (on X-axis) and non-robust feature (on Y-axis, Figure \ref{fig:toy}) in order to investigate whether the latent adversarial perturbation helps filtering out the non-robust feature.

A simple neural network (\(L=2\)) is implemented where the adversarial perturbation is injected to each input layer and latent layer, i.e., \(K=\{0, 1\}\). Then we compare the decision boundary of binary classifiers obtained through standard training, FGSM AT, and \textbf{SLAT} (\(\eta_0=0.1\) for both FGSM AT and \textbf{SLAT}). Under the limited adversarial budgets, while FGSM AT still heavily relies on the predictive but non-robust feature (Figure \ref{fig:bdry_FGSM}), \textbf{SLAT} learns to select a more robust feature (Figure \ref{fig:bdry_advGNI}). It implies that the latent adversarial perturbation collaborates with the input adversarial perturbation, so that the effect of adversarial training can be amplified.

\subsection{Comparison of robustness}

\textbf{Datasets.}  To further investigate the effect of latent adversarial perturbation on adversarial robustness, we compare the adversarial robustness of several models on CIFAR-10, CIFAR-100 and Tiny ImageNet. For CIFAR-10 and CIFAR-100, all the images are randomly cropped into \(32 \times 32\) with padding size \(4\) following \cite{wong2020fast, andriushchenko2020understanding}. Images in Tiny ImageNet are similarly cropped into \(64 \times 64\).

\textbf{Experimental setup.}  For an adversarial example, \(\ell_\infty\)-perturbation is used with radius \(\eta_0 = 8/255\). For a fair comparison, we reproduced all the other baseline results using the same back-bone architecture and the optimization settings. Specifically, we validate several AT methods using Wide ResNet 28-10 \cite{zagoruyko2016wide}, and cyclic learning rates \cite{smith2019super} with the SGD optimizer following the setup of \cite{andriushchenko2020understanding}. 

The latent adversarial perturbation is injected into three layers (\(K\)), including input layer and last layers in each group \(conv1, conv2\) \cite{zagoruyko2016wide}. We use \(\eta_k=8/255\) for every \(k \in K\). We evaluate the adversarial robustness of several models using PGD-50-10 attack \cite{madry2017towards}, i.e., with 50 iterations and 10 restarts, and AutoAttack \cite{croce2020reliable} which is the ensemble of two extended PGD attacks, a white-box FAB-attack \cite{croce2020minimally}, and the black-box Square Attack \cite{andriushchenko2020square}, for verifying the absence of gradient masking \cite{athalye2018obfuscated}. Every experiments are run on a single GeForce Titan X. The details regarding the simulation settings are presented in the supplementary material.

\textbf{Baselines.} We compare our method with the following state-of-the-art (fast) adversarial training methods: PGD AT \cite{madry2017towards}, FGSM-GA (FGSM AT with gradient alignment, \cite{andriushchenko2020understanding}), YOPO \cite{zhang2019you}, Free-AT \cite{shafahi2019adversarial}, FGSM AT \cite{goodfellow2014explaining}, FGSM-RS (FGSM AT with random step, \cite{wong2020fast}), and FGSM-CKPT (FGSM AT with checkpoints, \cite{kim2020understanding}). Every model is trained based on the reported hyperparameter configurations of original papers except FGSM-GA which experiences catastrophic overfitting with the reported gradient alignment parameter \(\lambda=0.2\). We therefore tested with the parameter \(\lambda \in \{0.2, 0.3, 0.5, 1.0, 2.0\}\) where \(\lambda=1.0\) is selected.

\textbf{Results.} The clean and robust accuracy for the CIFAR-10, CIFAR-100 and Tiny ImageNet datasets are summarized in Table \ref{table:CIFAR-10}. We found that \textbf{SLAT} reliably outperforms most of the accelerated adversarial training methods, including YOPO \cite{zhang2019you}, Free-AT \cite{shafahi2019adversarial}, FGSM-RS, and FGSM-CKPT \cite{kim2020understanding}, with respect to robust accuracy against PGD attack and AutoAttack \cite{croce2020reliable}, on every experiments. Note that most of the FGSM-based accelerated AT methods experienced a catastrophic overfitting during the training process on specific datasets, e.g., FGSM on CIFAR-10, CIFAR-100; FGSM-CKPT on CIFAR-100, Tiny ImageNet; FGSM-RS on CIFAR-10, CIFAR-100. Although FGSM-GA \cite{andriushchenko2020understanding} and PGD-7 tend to exhibit better performance than \textbf{SLAT}, both methods are much slower than the other adversarial training methods. Moreover, \textbf{SLAT} achieves a higher standard accuracy than both methods on CIFAR-10. 

\subsection{Quantitative and Qualitative analysis}

\begin{figure}[htbp]
\centering
\begin{subfigure}[c]{0.23\textwidth}
\includegraphics[width=\textwidth]{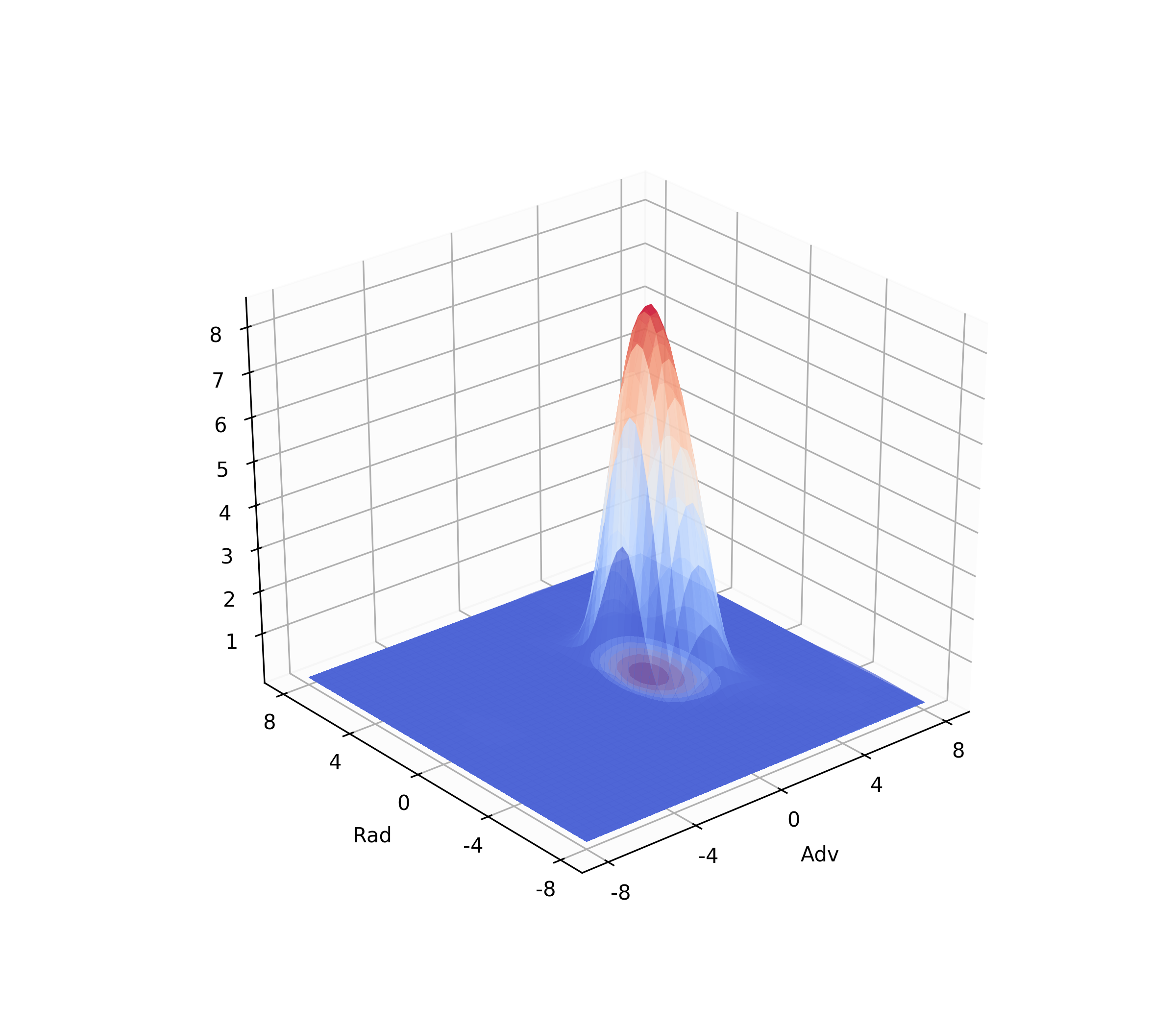} 
\caption{FGSM}\label{fig:loss_FGSM}
\end{subfigure}
\begin{subfigure}[c]{0.23\textwidth}
\includegraphics[width=\textwidth]{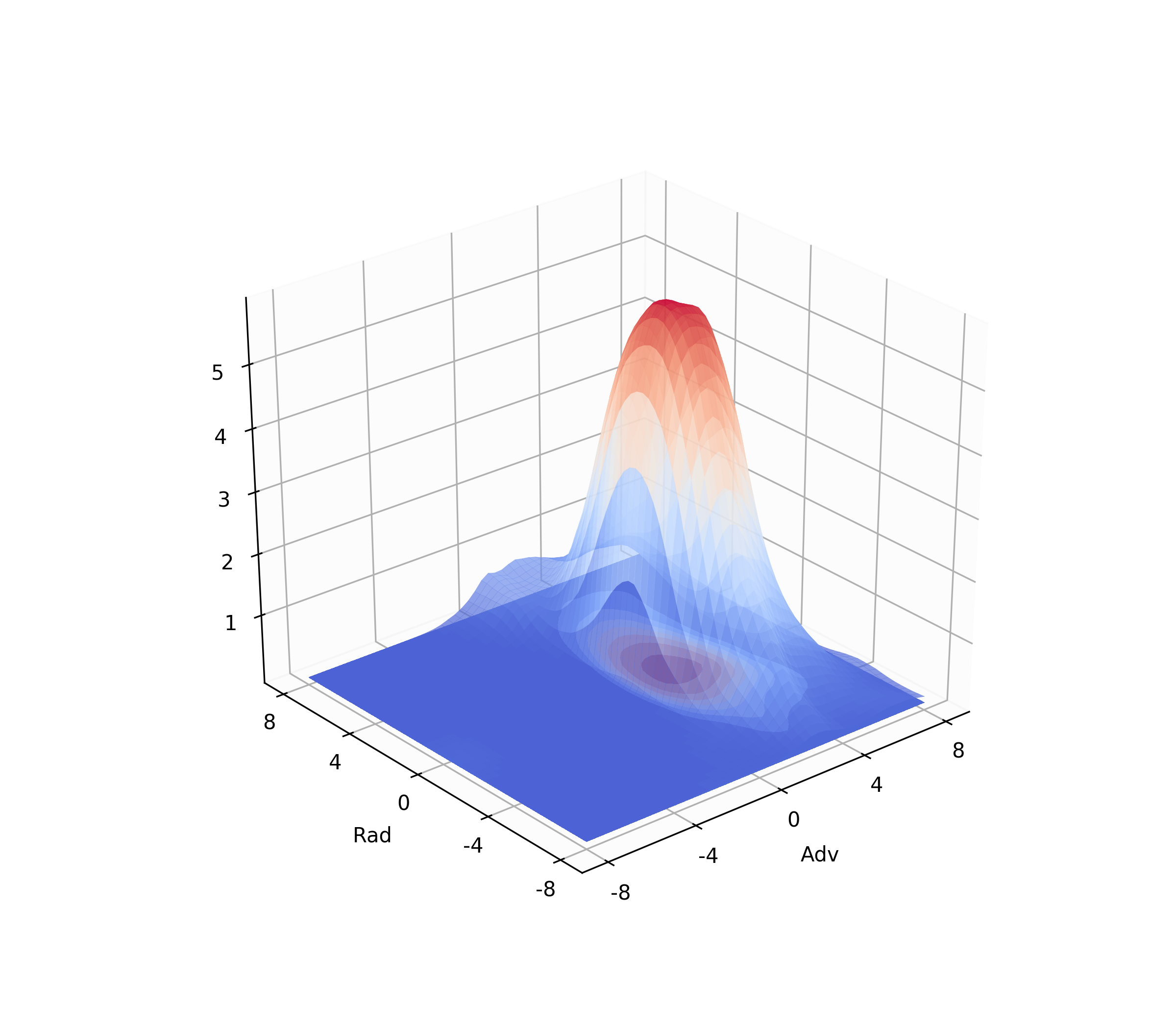}
\caption{FGSM-RS}\label{fig:loss_FGSM_RS}
\end{subfigure}
\begin{subfigure}[c]{0.23\textwidth}
\includegraphics[width=\textwidth]{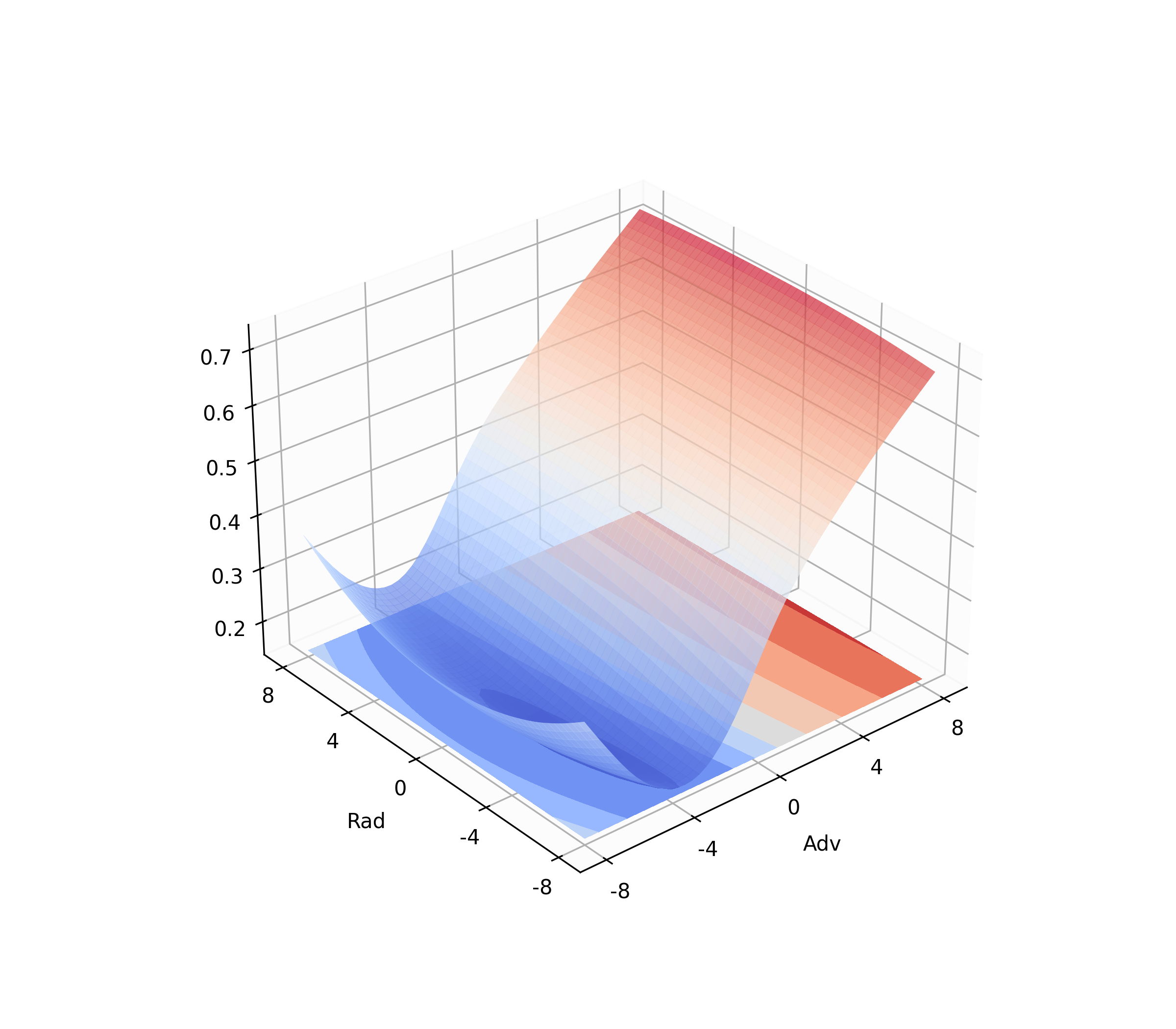}  
\caption{FGSM-CKPT}\label{fig:loss_FGSM_ckpt}
\end{subfigure}
\begin{subfigure}[c]{0.23\textwidth}
\includegraphics[width=\textwidth]{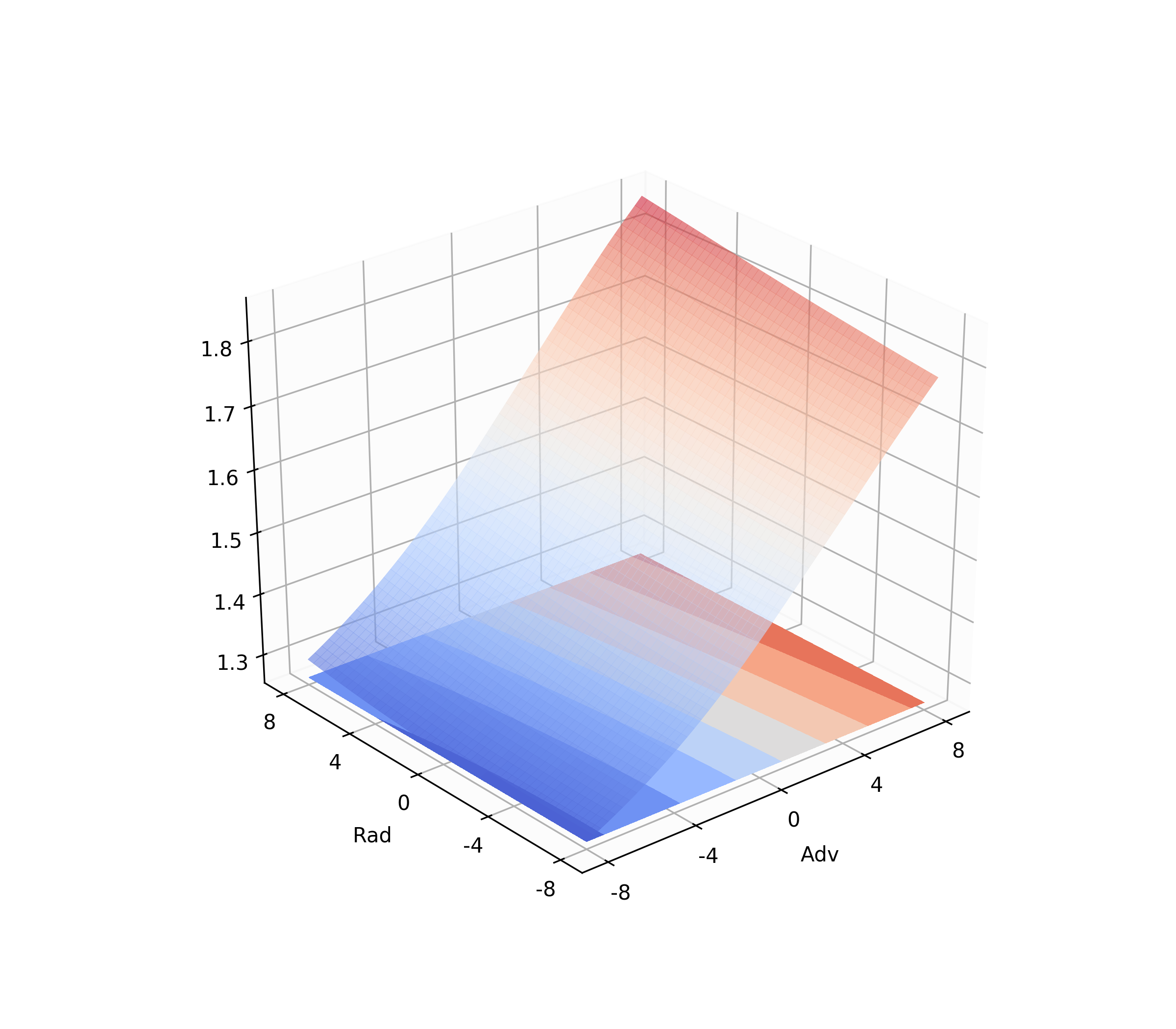} 
\caption{\textbf{SLAT}}\label{fig:loss_SSH_AT}
\end{subfigure}
\caption{Visualization of loss landscape on CIFAR-10 for various models. We plot the softmax cross entropy loss projected on adversarial direction and random (Rademacher) direction with \(\eta_0=8/255\) radius.} \label{fig:loss_landscape}
\end{figure}
\begin{figure}[htbp]
\centering

\begin{subfigure}[c]{0.23\textwidth}
\includegraphics[width=\textwidth]{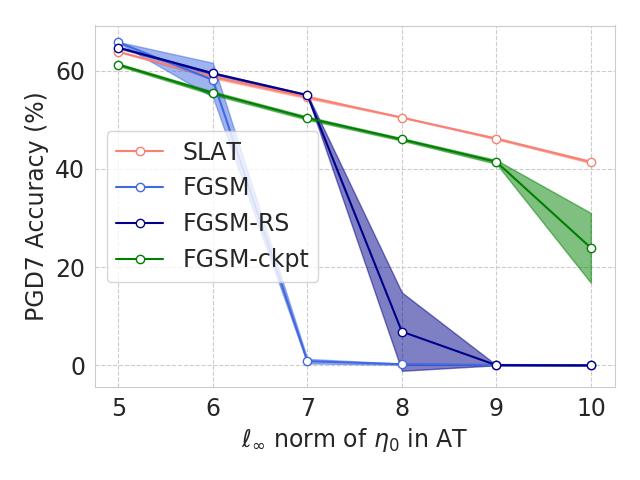}
\caption{Reliability}\label{fig:eps_acc}
\end{subfigure}
\begin{subfigure}[c]{0.23\textwidth}
\includegraphics[width=\textwidth]{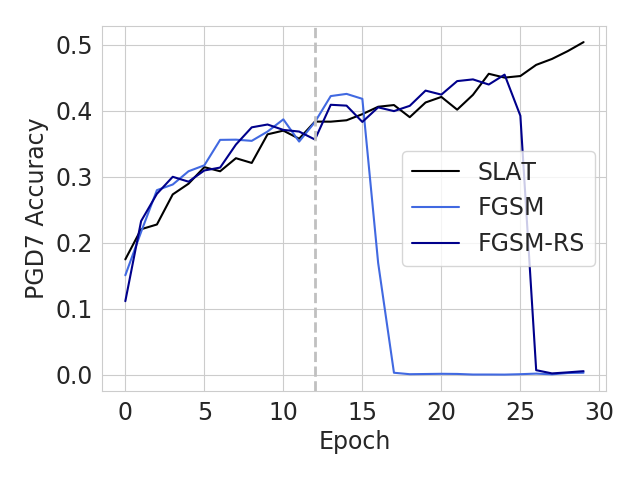}
\caption{Robust accuracy}\label{fig:adv_acc}
\end{subfigure}
\begin{subfigure}[c]{0.23\textwidth}
\includegraphics[width=\textwidth]{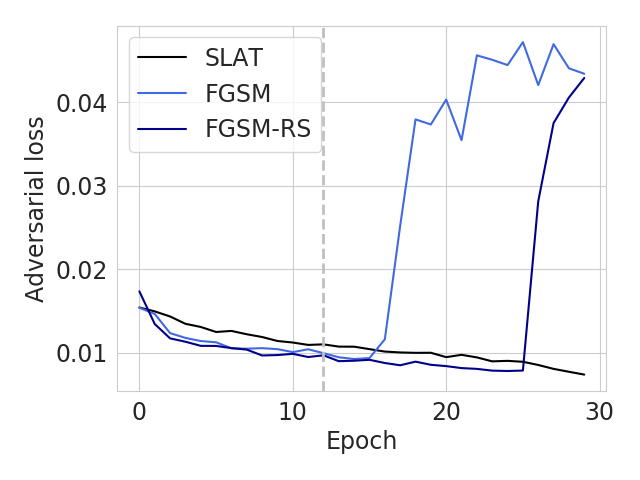}
\caption{Adversarial loss}\label{fig:adv_loss}
\end{subfigure}
\begin{subfigure}[c]{0.23\textwidth}
\includegraphics[width=\textwidth]{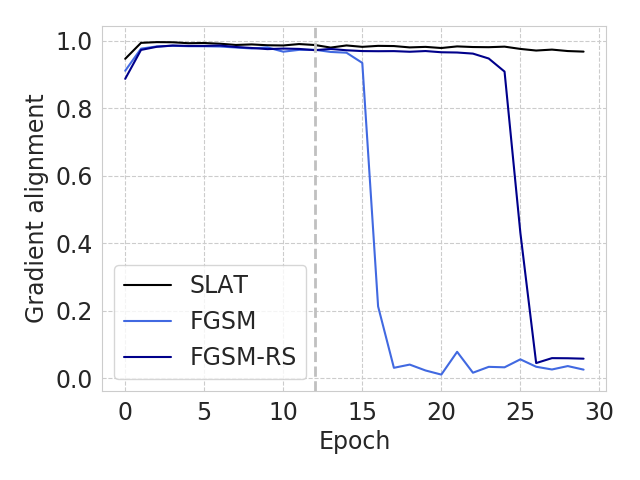}  
\caption{Gradient alignment}\label{fig:adv_cos}
\end{subfigure}
\caption{\textbf{(a)} Robust accuracy (\(\%\)) of various AT methods with different \(\ell_\infty\) radius on CIFAR-10. The results are averaged on 4 different random seeds. \textbf{(b)\(\sim\)(d)} Demonstrating the relationship between catastrophic overfitting and local linearity on CIFAR-10. Gray dotted line indicates the epoch at maximum learning rate.} \label{fig:adv_linearity}
\end{figure}

\textbf{Visualizing loss landscape.}   To fully justify the association between latent hidden perturbation and local linearity, we visualize the loss landscape of several models. Figure \ref{fig:loss_SSH_AT} shows that the loss of \textbf{SLAT} in \(\ell_\infty\) ball is almost perfectly in linear with the adversarial direction, thus qualitatively proving that the local linearity is recovered by adopting latent perturbation. Furthermore, Figure \ref{fig:loss_FGSM}, \ref{fig:loss_FGSM_RS} shows that the loss landscape of FGSM and FGSM-RS are highly non-linear in the vicinity of training examples. Note that the adversarial loss of FGSM-RS given adversary with maximum step size is relatively lower than the other well-trained methods (Figure \ref{fig:loss_FGSM_ckpt}, \ref{fig:loss_SSH_AT}). Due to the catastrophic overfitting problem, both methods are still vulnerable to other smaller perturbations. Although FGSM-CKPT \cite{kim2020understanding} performs better than FGSM-RS, the obtained loss landscape is not perfectly linear as \textbf{SLAT} in adversarial direction.

\textbf{Evaluating adversarial robustness with different \(\ell_\infty\) radius.} To investigate the contribution of latent adversarial perturbation on reliability, we compare the robustness of FGSM-based adversarial training methods with different \(\ell_\infty\)-radius (Figure \ref{fig:eps_acc}). The results illustrate that the latent adversarial perturbation prevents the model from losing its robustness quickly when the \(\ell_\infty\)-radius increases. Note that the FGSM-based single-step adversarial training methods (FGSM, FGSM-RS, FGSM-CKPT) are relatively sensitive to the \(\ell_\infty\)-radius, potentially due to the lack of regularization for a better linear approximation.

\textbf{Measuring local linearity.} To specify the connection between catastrophic over-fitting and local linearity, we measure the adversarial robustness and gradient alignment as done in \cite{andriushchenko2020square} (Figure \ref{fig:adv_acc}, \ref{fig:adv_loss}, \ref{fig:adv_cos}). Gradient alignment is measured as the cosine similarity between input gradients computed with original and randomly perturbed sample. We figured out that the gradient alignment suddenly drops as the model loses its adversarial robustness. While FGSM or FGSM-RS may rely on the early-stopping technique \cite{wong2020fast} to empirically prevent catastrophic overfitting, it is not sufficient to outperform \textbf{SLAT} (Figure \ref{fig:adv_acc}).

\begin{figure*}[ht]
    \centering
    \includegraphics[width=\textwidth]{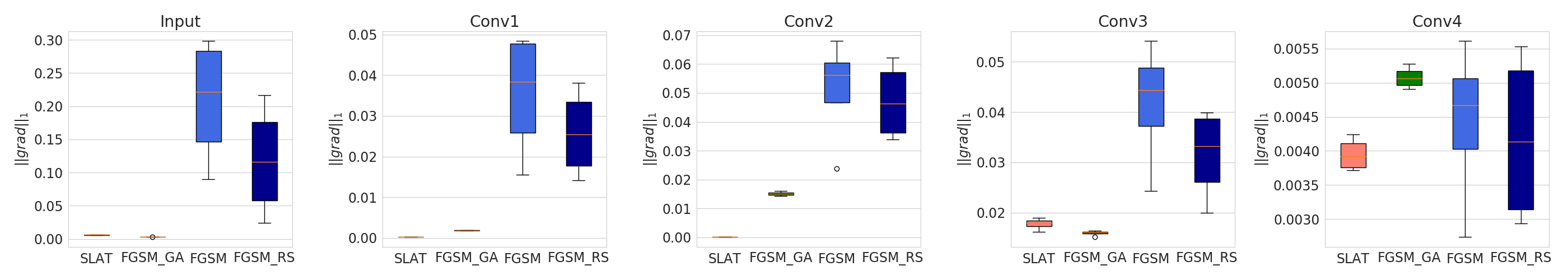} 
    \caption{\(\ell_1\) norm of gradients on CIFAR-10 after training. The results are averaged on 4 different random seeds.}\label{fig:grad_norm}
\end{figure*}

\textbf{Measuring \(\ell_1\) norm of feature gradients.} With respect to the analysis in the section \ref{sec:latent}, we compared the \(\ell_1\) norm of the stochastic gradients computed with different methods. After training, we computed the \(\ell_1\) norm of gradients for every representation \(h_l(x)\) for all \(l \in K\). Figure \ref{fig:grad_norm} shows that \textbf{SLAT} yields lower norm of the gradients compared to the other methods on layers injected with latent perturbation (\textit{Input, Conv1, Conv2}). It is potentially due to the synergistic regularization by latent adversarial perturbation. 

\textbf{Ablation study on latent adversarial perturbation.} To quantify the extent of performance improvement achieved by the latent adversarial perturbation, we compared the standard and robust accuracy of the different versions of \textbf{SLAT} (Table \ref{table:ablation}). When the latent adversarial perturbation is excluded, the proposed method reduces to the conventional FGSM AT. To consider the effect of step size reduction \cite{andriushchenko2020understanding}, we also measured the performance of the version of FGSM AT with a reduced step size. Table \ref{table:ablation} shows that the vanilla FGSM AT is not sufficient to prevent the catastrophic overfitting problem regardless of step size. Moreover, we found that latent adversarial perturbations prevent the catastrophic overfitting of FGSM-RS \cite{wong2020fast}. This result indicates that the latent adversarial perturbation plays a important role in enhancing the reliability of adversarial training. Although the latent adversarial perturbations improve the adversarial robustness of FGSM-RS to some extent, it does not outperform \textbf{SLAT}. It is because the latent adversarial perturbation was derived from the original sample rather than the randomly perturbed sample as in (\ref{eq:latent_adv_solution}). 

\begin{table}[htbp]
\caption{Ablation study on latent adversarial perturbation with \(\eta_0 = 8/255\) (CIFAR-10).} 
\label{table:ablation}
\centering
\begin{tabular}{c c c}
\toprule
Method & {Standard} & {PGD-50-10} \\
\midrule
\specialcell{FGSM \\ (step size\(=8/255\))} & {87.42} & {0.01} \\
\midrule
\specialcell{FGSM \\ (step size\(=0.9*8/255\))} & {90.87} & {3.09} \\
\midrule
\specialcell{FGSM-RS \\ (w/ latent perturbation)} & {82.87} & {44.95} \\
\midrule
\textbf{SLAT} & {85.91} & {47.06} \\
\bottomrule
\end{tabular}
\end{table}

Moreover, to understand the effect of layer depth on latent adversarial training, we compared the performance of \textbf{SLAT} with different subsets of layer indexes \(K\). Following \cite{zagoruyko2016wide}, latent adversarial perturbation was added to the last layer of some selected blocks among \textit{conv1, conv2, conv3, conv4} of Wide ResNet 28-10. The input layer was included equally in \(K\) for all experiments. We experimentally found that the robustness and standard accuracy decreased significantly when the latent adversarial perturbations were injected into deeper layers (Table \ref{table:layer_depth}). This implies that simply injecting latent adversarial perturbation to arbitrarily many hidden layers is not necessarily be effective and may over-regularize the networks. These findings provide us an interesting conjecture that the recovery of local linearity should be primarily confined to early sub-networks that recursively include other deeper sub-networks. The theoretical analysis is left for future research.

\begin{table}[htbp]
\caption{Analysis of adversarial robustness and standard accuracy based on layer depth change (CIFAR-10).} 
\label{table:layer_depth}
\centering
\begin{tabular}{c c c c}
\toprule
{Layers} & {Standard} & {PGD-50-10} & {AutoAttack} \\
\midrule
\specialcell{\textit{conv1, conv2} \\ (\textbf{SLAT})} & {85.91} & {47.06} & {44.62} \\
\midrule
\textit{conv3, conv4} & {84.60} & {0.00} & {0.00} \\
\midrule
\specialcell{\textit{conv1, conv2}, \\ \textit{conv3, conv4}} & {81.41} & {47.3} & {43.58} \\
\bottomrule
\end{tabular}
\end{table}

\begin{figure}[htbp]
\centering

\begin{subfigure}[c]{0.23\textwidth}
\includegraphics[width=\textwidth]{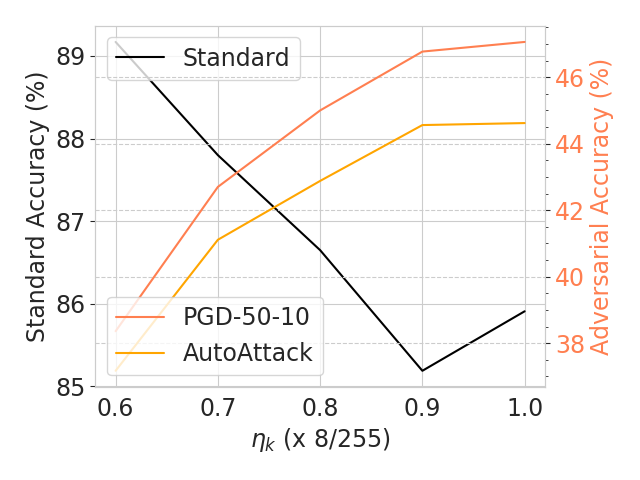}
\caption{Hyperparameter analysis}\label{fig:hyper_sensitivity}
\end{subfigure}
\begin{subfigure}[c]{0.23\textwidth}
\includegraphics[width=\textwidth]{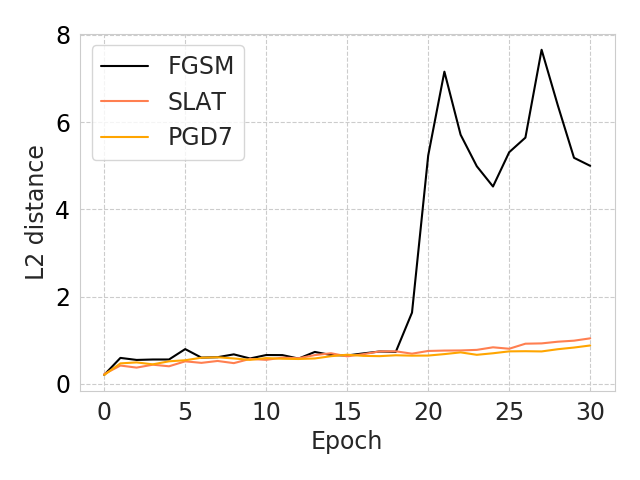}
\caption{\(\ell_2\)-distance}\label{fig:gradient_masking}
\end{subfigure}
\caption{\textbf{(a)} Impact of \(\eta_k\) on robustness. \textbf{(b)} Average \(\ell_2\) distance between logits of FGSM and R+FGSM \cite{tramer2017ensemble} adversary (CIFAR-10).}
\end{figure}

We additionally conducted the analysis on hyperparameter sensitivity with the adversarial step size \(\eta_k\) (Figure \ref{fig:hyper_sensitivity}). Standard and adversarial accuracy are measured on CIFAR-10, where the step size \(\eta_k\) varies from \(0.6*8/255\) to \(1.0*8/255\). The robust accuracy is high when \(\eta_k=8/255\). Thus, we fix \(\eta_k\) as \(8/255\) for every dataset and layer \(k \in K\). It may work better to fine-tune \(\eta_k\) differently depending on \(k\). From our preliminary analyses, the unified \(\eta_k\) for the latent representations worked sufficiently well that we did not feel the need to fine-tune \(\eta_k\) for every layer. Moreover, while the trade-off between robustness and accuracy is observed, the standard accuracy for \(\eta_k=8/255\) is still superior than that of FGSM-GA or PGD-7 (Table \ref{table:CIFAR-10}).

\textbf{Verifying absence of gradient masking.} Although many certified adversarial defense methods have reported substantial advances in adversarial robustness, recent works \cite{papernot2017practical, engstrom2018evaluating, athalye2018obfuscated} report that the gradient masking may present a false sense of security, i.e., making it difficult to generate adversary using gradient methods by obfuscating gradients, while the adversary still potentially exists. Besides the quantitative (Autoattack in Table \ref{table:CIFAR-10}) and qualitative (Figure \ref{fig:loss_landscape}, \cite{qin2019adversarial, shafahi2019adversarial}) evidences, we additionally measure the \(\ell_2\) distance between logits of FGSM and R+FGSM \cite{tramer2017ensemble} adversary to confirm the absence of gradient masking following \cite{carlini2019evaluating, vivek2019regularizer, tramer2017ensemble}. If the model exhibits gradient masking, the distance between logits of FGSM and R+FGSM increases due to the sharp curvature of loss landscape in vicinity of training examples. Figure \ref{fig:gradient_masking} shows that the distance of proposed method is relatively low compared to FGSM AT during the training process, thus proving that the proposed method does not rely on gradient masking.

\section{Conclusion}
In this study, we demonstrate that the latent adversarial perturbation may provide a novel breakthrough for the efficient AT. The proposed framework allows us to compensate for the local linearity without sacrificing training time. Further, we establish a bridge between latent adversarial perturbation and adversarial loss minimization. It enables us to learn adversarially robust model in a more reliable manner, compared to the recent fast adversarial training method \cite{wong2020fast} which lacks any form of regularization. By running simulations on various benchmark datasets, we illustrate that our model significantly outperforms state-of-the-art accelerated adversarial training methods. The proposed method is fully-architecture agnostic, has only a few free parameters to tune, and is potentially compatible with many other AT methods. We believe that expanding the proposed framework beyond
single-step AT will be an interesting future work. 

\subsubsection*{Acknowledgments}
This work was supported by Institute for Information \& Communications Technology Promotion (IITP) grant funded by the Korea government (No. 2017-0-00451) (No.2019-0-01371, Development of brain-inspired AI with human-like intelligence), National Research Foundation of Korea (NRF) grant funded by the Korea government (MSIT) (NRF-2019M3E5D2A01066267), and Samsung Research Funding Center of Samsung Electronics under Project Number SRFC-TC1603-06.


{\small
\bibliographystyle{ieee_fullname}
\bibliography{egbib}
}

\clearpage

\appendix

This supplementary material is organized as follows. In section \ref{sec:proof}, we present the proof for Proposition \ref{pro:accum_perturb}, which is obtained by modifying \cite{camuto2020explicit}, for completeness. Optimization setting and hyperparameter configurations are presented in section \ref{sec:setup}.

\section{Proof}
\label{sec:proof}

\begin{custompro}{1}
\label{pro:accum_perturb}
Consider an \(L\) layer neural network, with the latent adversarial perturbations \(\delta_k(x)\) being applied at each layer \(k \in K\). Assuming the Hessians, of the form \(\nabla^2 h_l(x)|_{h_m(x)}\) where \(l, m\) are the index over layers, are finite. Then the perturbation accumulated at the layer \(L-1\), \(\hat{\delta}_{\ell-1}(x)\), is approximated by:
\begin{equation}
\hat{\delta}_{\ell-1}(x) = \sum_{k \in K} \mathbf{J}_k(x) \delta_k(x) + O(\gamma),
\end{equation}

where \(\mathbf{J}_k(x) \in \mathbb{R}^{N_{L-1} \times N_k}\) represents each layer's Jacobian; \(\mathbf{J}_k(x)_{i,j} = \frac{\partial h_{L-1}(x)_i}{\partial h_k(x)_j}\), given the number of neurons in layer \(L-1\) and \(k\) as \(N_{L-1}\) and \(N_k\), respectively. \(O(\gamma)\) represents higher order terms in \(\pmb{\delta}\) that tend to zero in the limit of small perturbation.
\end{custompro}

\begin{proof}
Starting with layer 0 as the input layer, the accumulated perturbation on a layer \(L-1\) can be approximated through recursion. Following the conventional adversarial training, suppose that the first layer index \(0\) is included in the set \(K\). At layer 0, we apply Taylor's theorem on \(h_1(x+\delta_0(x))\) around the original input \(x\). If we assume that all values in Hessian of \(h_1(x)\) is finite, i.e., \(|\partial^2 h_1(x)_i / \partial x_j x_k | < \infty, \forall i,j,k\), the following approximation holds:
\begin{equation}
    \label{eq:h1_approx}
    h_1(x+\delta_0(x)) = h_1(x) + \frac{\partial h_1(x)}{\partial x} \delta_0(x) + O(\kappa_0),
\end{equation}
where \(O(\kappa_0)\) represents asymptotically dominated higher order terms given the small perturbation. By accommodating \(L=2\) as a special case, we obtain the accumulated noise \(\delta_{\ell-1}=\frac{\partial h_1(x)}{\partial x} \delta_0(x) + O(\kappa_0)\). Note that (\ref{eq:h1_approx}) can be generalized with an arbitrary layer index \(k+1\) and perturbation \(\delta_k(x)\). 

Repeating this process for each layer \(k\in K\) recursively, and assuming that all Hessians of the form \(\nabla^2 h_l(x)|_{h_m(x)}, \forall m<l\) are finite, we obtain the accumulated perturbation for a layer \(L-1\) as follows:
\begin{equation}
    \label{eq:accum_noise}
    \hat{\delta}_{\ell-1}(x) = \sum_{k \in K} \frac{\partial h_{L-1}(x)}{\partial h_k(x)} \delta_k(x) + O(\gamma),
\end{equation}
where \(O(\gamma)\) represents asymptotically dominated higher order terms as the perturbation \(\delta_k(x), \forall k \in K\) is sufficiently small. Denoting \(\frac{\partial h_{L-1}(x)}{\partial h_k(x)}\) as the Jacobian \(\mathbf{J}_k(x) \in \mathbb{R}^{N_{L-1} \times N_k}\) completes the proof.
\end{proof}

\section{Experimental setup}
\label{sec:setup}
We provide extended details about simulation settings for completeness. For a fair comparison, we reproduced all the other baseline results using the same back-bone architecture and the optimization settings. Every method is trained for \(30\) epochs except Free-AT \cite{shafahi2019adversarial} which is trained for \(72\) epochs to get results comparable to the other methods. Following the setup of \cite{andriushchenko2020understanding}, we use cyclic learning rates \cite{smith2019super} with the SGD optimizer with momentum 0.9 and weight decay \(5*10^{-4}\). Specifically, the learning rate increases linearly from 0 to 0.2 in first 12 epochs, and then decreases linearly to 0 in left 18 epochs. We use a batch size of 128 for CIFAR-10 and CIFAR-100 experiments. For the Tiny ImageNet experiments, we use a batch size of 64 to reduce the memory consumption. 

For FGSM-RS \cite{wong2020fast} on every dataset, we use a step size \(\alpha=1.25\eta_0\) following the recommendation of authors. We succeeded in reproducing the robust accuracy of FGSM-RS against PGD-50-10 attack using the experimental setup reported in \cite{wong2020fast}, but found that catastrophic overfitting occurs when the epoch was increased to 30. For PGD-7 AT, we use a step size \(\alpha=2\eta_0/10\) for generating a 7-step PGD adversarial attack sample. 

\end{document}